\documentclass[table]{scrartcl}
\usepackage[utf8]{inputenc}
\usepackage{amsmath}
\usepackage[english]{babel}
\usepackage[margin=1in]{geometry}
\usepackage{setspace}
\usepackage{natbib}
\usepackage{graphicx}
\usepackage{tikz}
\usepackage{scrextend}
\usetikzlibrary{shapes,arrows}
\usepackage{afterpage}
\usetikzlibrary{shadows}
\usetikzlibrary{shapes, shapes.geometric}
\usepackage{caption}
\usepackage{hyperref}
\setcitestyle{authoryear,open={(},close={)}}
\usepackage{authblk}
\usepackage[utf8]{inputenc}
\usepackage{subfiles}
\usepackage{multirow}
\newtheorem{theorem}{Definition}
\definecolor{PANDarkGray}{RGB}{153,153,153}

\usepackage{pifont}
\usepackage{amssymb}
\newcommand{\xmark}{\ding{55}}

\usepackage[breakable]{tcolorbox}
\usepackage{parskip} 
    
\usepackage{listings}
\usepackage{color}

\definecolor{dkgreen}{rgb}{0,0.6,0}
\definecolor{gray}{rgb}{0.5,0.5,0.5}
\definecolor{mauve}{rgb}{0.58,0,0.82}

\title{Stance Detection: A Practical Guide to Classifying Political Beliefs in Text}
\author{Michael Burnham}
\date{November, 2023}
\affil[]{Department of Political Science and Center for Social Data Analytics\\ Penn State University}
\begin{document}

\maketitle
\begin{abstract}
\noindent Stance detection is identifying expressed beliefs in a document. While researchers widely use sentiment analysis for this, recent research demonstrates that sentiment and stance are distinct. This paper advances text analysis methods by precisely defining stance detection and presenting three distinct approaches: supervised classification, natural language inference, and in-context learning with generative language models. I discuss how document context and trade-offs between resources and workload should inform your methods. For all three approaches I provide guidance on application and validation techniques, as well as coding tutorials for implementation. Finally, I demonstrate how newer classification approaches can replicate supervised classifiers.
\end{abstract}
\doublespacing
\section{Introduction}

Stance detection is a common tasks for political scientists conducting text analysis.\footnote{Coding tutorials demonstrating methods presented are included in the appendix} It may include identifying support or opposition towards electoral candidates \citep{barbera2016less} or social movements \citep{felmlee2020geography}. It may involve identifying attitudes towards political \citep{bor2023discriminatory}, religious \citep{terman2017islamophobia} or racial \citep{lee2022racism} groups. Or, researchers may measure positions on factual matters such as whether or not COVID-19 poses a public health risk \citep{block2022perceived} or other types of misinformation \citep{osmundsen2021partisan}.

To measure stance, social scientists widely rely on sentiment analysis -- the identification of positive or negative emotion in text \citep{liu2010sentiment, hutto2014vader}. However, recent research shows sentiment is often uncorrelated with opinion \citep{aldayel2019assessing, bestvater2022sentiment}. As \citet{aldayel2021stance} argue, sentiment is a dimension distinct from stance.

This paper builds on \citet{aldayel2021stance} and \citet{bestvater2022sentiment} who conclude that stance detection is a classification task and recommend training supervised classifiers. Although text classification is well established in political science, opinion classification is particularly difficult and warrants special considerations \citep{aldayel2021stance, aldayel2019assessing}. Further, in recent years, language models introduced new approaches capable of labeling documents without supervised training. Thus, guidance beyond training a classifier is merited. 

I advance theory and methods in stance detection in three ways: First I provide a precise definition of stance detection. Second, I provide an overview of three classification paradigms: Supervised classification, which trains language models for a specific task; natural language inference (NLI), which uses language models pre-trained as ``universal'' classifiers; and in-context learning, which uses generative models, like GPT-4, to generate labels based on plain language descriptions of the task. Finally, I provide practical guidance on selecting and implementing an appropriate approach. This article presents a conceptual overview of the motivations and implementations for each paradigm. For guidance on technical implementation I include coding tutorials in the replication and online materials.

I particularly emphasize NLI classifiers. While supervised methods are well established and generative models have attracted significant attention from political scientists \citep[e.g.][]{tornberg2023use, gilardi2023chatgpt}, NLI classifiers are largely absent from the literature despite their advantages: classification that scales to large data sets, reproduces well, and does not require training. In the final section, I use an NLI classifier to replicate the findings of \citet{block2022perceived} -- which used a supervised classifier to label stances towards COVID-19 mitigation policies.

\label{sec:intro}

\section{Stance and Stance Detection}

\subsection{What is Stance?}
Stance is an individual's ``attitudes, feelings, judgments, or commitment'' to a proposition \citep{biber1988adverbial}. In the recent past, stance detection was synonymous with sentiment analysis that measures polarity of documents on a positive or negative scale \citep{stine2019sentiment}. However, this introduces significant room for measurement error. The position a document expresses is often different from the sentiment used to express it. For example: ``So excited to see Trump out of the White House!'' expresses a negative stance about Trump with positive sentiment. Sentiment can potentially inform intensity of stance, but they are distinct dimensions requiring independent measurement \citep{aldayel2019assessing}.

\citet{bestvater2022sentiment} and \citet{sasaki2016stance} propose targeted sentiment (i.e. sentiment towards Trump rather than general document sentiment) as an operationalization for stance. However, positive and negative sentiment is one dimension of ``attitudes, feelings, judgments, or commitment.'' If stance is how individuals would answer a proposition, answers need not contain sentiment. The statement ``I'm voting for Joe Biden'' is a positive response to the proposition ``Are you voting for Joe Biden?'' but lacks sentiment cues. It may be the dejected expression of progressive liberal who sees Biden as the least-bad option (negative sentiment), or an enthusiastic supporter (positive sentiment). Characterizing the statement as positive sentiment conflates sentiment and stance dimensions and excludes the former possibility. Sentiment should not be both how someone would answer a proposition and how they express that answer. 

Accordingly, I use two simple definitions consistent with the broader literature and treat sentiment and stance independently:
\begin{theorem}
    Stance: How an individual would answer a proposition.
\end{theorem}
\begin{theorem}
    Sentiment: Positive or negative emotional valence.
\end{theorem}
Both sentiment and stance may be directed towards individuals, groups, policy, etc. but are distinct. 

\subsection{What is Stance Detection?}
Recent literature operationalizes stance detection as entailment classification \citep{aldayel2021stance}. Introduced by \citet{dagan2005pascal}, textual entailment is a directional relationship between two documents called the text (\textit{T}) and the hypothesis (\textit{H}). 
\begin{theorem}
Textual Entailment: Text sample \textit{T} entails hypothesis \textit{H} when a human reading \textit{T} would infer that \textit{H} is most likely true.
\end{theorem} Thus, stance detection has two components: First, the proposition on which an author may take a stance (the hypothesis) and second, the text from which a stance is inferred. For example, if the following tweet from President Trump is paired with the following hypothesis:\\
\indent \textbf{T}: \textit{It's freezing and snowing in New York -- we need global warming!}\\
\indent \textbf{H}: \textit{Trump supports global warming.}\\
\noindent we would conclude text \textit{T} entails hypothesis \textit{H}. Textual entailment does not test if a hypothesis is \textit{necessarily} true, but what humans would infer is most likely true. 

\citet{aldayel2021stance} adopt this definition and define stance detection such that ``a text entails a stance to a target if the stance to a target can be inferred from a given text''. Building on this, I adapt their definition:

\begin{theorem}
Stance Detection: Text sample \textit{T} entails stance \textit{S} to author \textit{A} when a human reading \textit{T} with context \textit{C} would infer that \textit{T} expresses support for \textit{S}.
\end{theorem}
To their definition, I specify that stance detection identifies the stance a document expresses. This excludes predicting topics like abortion policy preferences or party affiliation from documents not discussing those topics.

Most notably, I call attention to the significance of context. Context is information relevant to the stance of interest. Available context consists of information the document contains, and what the classifier knows. Missing context increases document ambiguity, which makes classification harder. Accurate stance detection minimizes ambiguity by ensuring sufficient information between the document's contents and the classifier's knowledge base. Consider the examples in Table \ref{tab:semeval} from a common stance detection dataset \citep{mohammad2016semeval}. Samples were manually labeled as supporting, opposing, or neutral towards Donald Trump. Each example highlights labeling inconsistencies resulting from missing context creating ambiguity. The first does not mention Donald Trump and seemingly reasons that someone espousing certain narratives about the United States founding probably dislikes Trump. Another labeler may think it is unrelated. Examples two and three express support for the Spanish news station Univision. Labelers from sample two seem aware that Trump had a disagreement with Univision and associate a pro-Univision stance with an anti-Trump stance. Sample three's labelers seemingly lack this context and assigned a different label. Accurate labeling can require time-sensitive knowledge of the context that produced a document. Unless the person or algorithm assigning labels has this information, accurate inference may be impossible.

\begin{table}[]
    \centering
    \begin{tabular}{p{.25cm}|p{6cm}|p{1.5cm}|p{5cm}}
        \hline
        \# & Text & Label & Notes \\
        \hline
        1 & @ABC Stupid is as stupid does! Showed his true colors; seems that he ignores that US was invaded, \& plundered, not discovered & Against & Based on external knowledge that this is about Trump, potentially inferred from stance about narrative of US founding \\
        
        2 & @peddoc63 @realDonaldTrump So I guess Univision is Fair \& Balanced. These are the people that R helping shape USA! & Against & Infers stance towards Trump based on positive stance towards Univision \\
        
        3 & Honestly I am gonna watch \#Univision so much more now, just to support the network against & Neutral & Expresses the same stance as example \#2, but the labelers apparently did not have the same contextual information when inferring stance \\        
        \hline
    \end{tabular}
    \caption{Text samples from the Semeval 2016 test data set \citep{mohammad2016semeval}.}
    \label{tab:semeval}
\end{table}


\subsection{Methodological Implications}
Given the above, there are three methodological implications: First, stance detection is a classification task. This is because stance detection is textual entailment, and textual entailment is classifying text as entailing, contradicting, or neutral. 

Second, stance is a response to a proposal -- the hypothesis \textit{H} in the definition of entailment. The job of the researcher is to define what the proposal is, how to determine if a document is relevant to a proposal, and what constitutes support or opposition to that proposal.

Finally, stances inferred from a document are context dependent. For accurate inference, researchers should try to control context by accounting for what information documents contain, and what information the classifier knows. In practical terms, this implies first making informed decisions about how documents are sampled or prepared, and second, choosing the right classification approach. The following section outlines both of these decisions to aid researchers in being more mindful of context and its effects.

\label{sec:lit}

\section{Controlling Context}

In this and the following section, I outline considerations for selecting and implementing a stance detection approach. To demonstrate the impact of these decisions, I use a dataset that is stance labeled for approval of President Trump. The data consists of both public tweets ($n = 1,135$) and a random sample of sentences containing the word ``Trump'' from congressional newsletters sent by the 117th congress ($n = 1,000$). The Twitter data was compiled from multiple stance detection projects and represents a variety of labeling approaches and document collection strategies. Details are listed in Appendix \ref{appendix: phrases}. Congressional newsletter sentences were collected from the DC Inbox project \citep{cormack2017dcinbox}. Data were coded twice by human annotators and discrepancies were adjudicated by a third annotator.

I use Matthew's Correlation Coefficient (MCC) as the primary performance metric due to its robustness relative to F1 and ROC AUC \citep{chicco2023matthews}. MCC ranges from -1 to 1 with 0 indicating no correlation between true class and estimated class.


\subsection{Document Preparation}
There are two primary ways to control context via the information documents contain: First, defining document relevancy and second, document pre-processing. Established methods for determining document relevancy during data collection include keyword matching, topic classification, or using content labels from an API. Stance detection does not necessitate a particular approach and the most appropriate will vary by project. However, researchers should be mindful that by setting inclusion criteria or sampling methods, researchers control context by setting parameters for what information documents contain. This, in turn, can inform your classification approach by determining what information your model should know. 

For example, if classifying stances on abortion I may include documents with the keyword ``Roe v. Wade''. Most documents will not explicitly state that support for Roe v. Wade equates with support for abortion rights. Accordingly, I should ensure my classification approach can make this association.

The second mechanism through which researchers can control context is document pre-processing. Most approaches to stance detection use language models that require minimal document pre-processing. Stop word removal, stemming, and other common procedures are generally detrimental for language models \citep{miyajiwala2022sensitivity}. Rather, the primary concern is how a ``document'' is defined. Short documents, like social media posts, may contain singular stance expressions. However, longer documents may contain multiple stance expressions across topics. In such circumstances it is advisable to segment documents and classify relevant segments to reduce noise.

Dividing documents into coherent segments is a challenging task. Existing methods include topic-based approaches \citep{riedl2012topictiling}, Bayesian models \citep{eisenstein2008bayesian}, and supervised classifiers \citep{koshorek2018text}. Here, I draw on \citet{barbera2021automated} who compared classification for individual sentences and article segments (roughly 5 sentences). They examined the trade-off between potentially discarding relevant text when classifying sentences, versus including irrelevant data when classifying sentence groups. They found that classifying at the sentence level reduced noise, but performance between approaches was similar. Thus, I advise simplicity. If there are natural segments in your documents (e.g. newspapers leads) multi-sentence segments may be appropriate. Otherwise, I recommend classifying sentences when segmentation is necessary. Sentence segmentation has many simple software implementations and the smaller document size of sentences can lead to significant efficiency gains.

\subsection{Model Selection}
With an understanding of the minimum context in your documents, you can make more informed decisions about selecting a classification method that will minimize ambiguity and bring requisite knowledge to the task. There are three approaches: supervised classifiers, NLI classifiers, and in-context classifiers. Each approach is further characterized by its training requirement and degree of control over the classifier's knowledge. Supervised classifiers require training data, but offer significant control over model knowledge. NLI and in-context classifiers, however, can label documents without training -- an ability known as ``zero-shot'' classification. NLI classifiers are more efficient but afford less control over model knowledge, while in-context classifiers provide significant knowledge control but require more compute. There is no best approach; rather, the goal is to optimize model performance, computational efficiency, and human workload. Table \ref{tab:class_types} overviews each method's strengths and weaknesses.

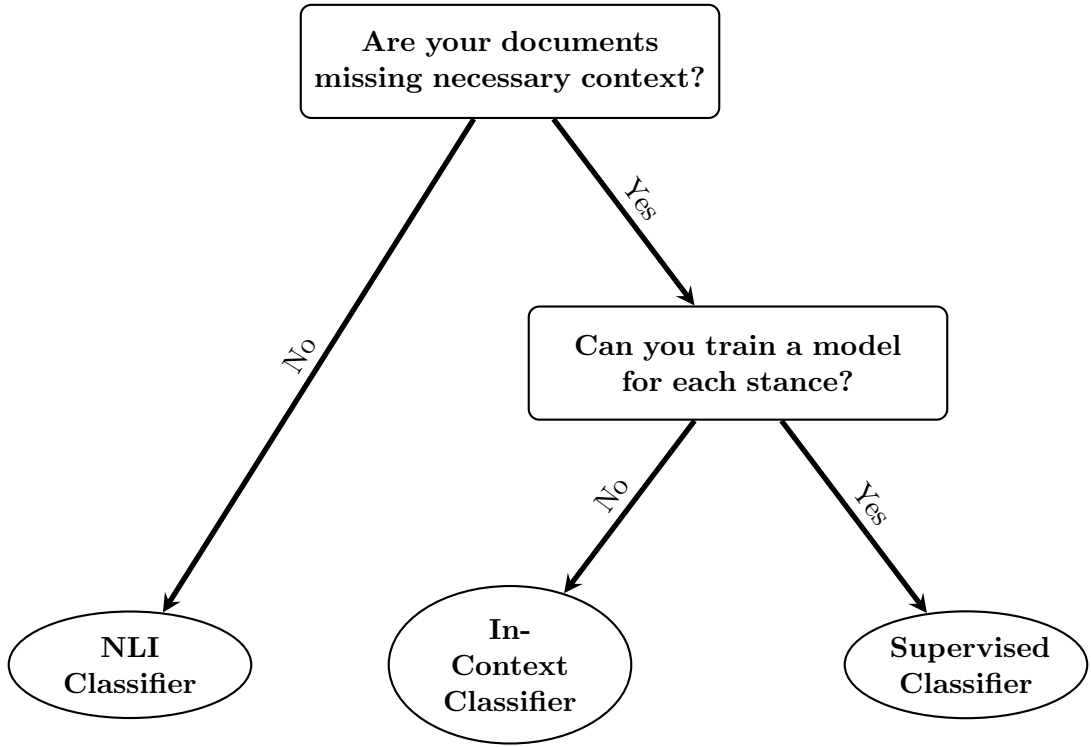
\begin{figure}
    \centering
        \begin{tikzpicture}[
            square/.style={rectangle, draw=black, thick, fill=white, minimum width=5.5cm, text width=5.25cm, minimum height=1.5cm, rounded corners},
            circle/.style={ellipse, draw=black, thick, fill=white, minimum width=3cm, text width=2cm, minimum height=1cm},
            arrow/.style={draw, thick,->,>=stealth, line width=2pt}
        ]
        
        \node[square, align = center] (scope) at  (-1,4) {\textbf{Are your documents\\ missing necessary context?}};
        \node[square, align = center] (context) at  (2,0) {\textbf{Can you train a model for each stance?}};
        \node[circle, align = center] (NLI) at  (-6,-4) {\textbf{NLI\\Classifier}};
        \node[circle, align = center] (incontext) at  (-1,-4) {\textbf{In-Context\\Classifier}};
        \node[circle, align = center] (supervised) at  (5,-4) {\textbf{Supervised\\Classifier}};
        
        \draw[arrow] (scope) -- node[sloped, anchor=center, above] {No} (NLI);
        \draw[arrow] (scope) -- node[sloped, anchor=center, above] {Yes}(context);
        \draw[arrow] (context) -- node[sloped, anchor=center, above] {No} (incontext);
        \draw[arrow] (context) -- node[sloped, anchor=center, above] {Yes}(supervised);
        
        \end{tikzpicture}
    \caption{The appropriate classification approach depends on the content of your documents and resource constraints.}
    \label{fig:flowchart}
\end{figure}

Figure \ref{fig:flowchart} presents a flow chart of key considerations, and the questions below expand on these considerations.

\begin{table}[h]
\centering
\resizebox{0.8\textwidth}{!}{%
\begin{tabular}{|l|c|c|c|c|}
\hline
\textbf{\begin{tabular}[c]{@{}l@{}}Classifier\end{tabular}} & \textbf{\begin{tabular}[c]{@{}c@{}}Requires\\ Training\end{tabular}} & \textbf{\begin{tabular}[c]{@{}c@{}}Compute \\ Requirements\end{tabular}} & \textbf{Reproducibility} & \textbf{Cost (Time/Money)} \\ \hline
Supervised                                                          & Yes                                                                  & CPU or single GPU                                                        & High  & High                    \\ \hline
NLI                                                                 & No                                                                   & CPU or single GPU                                                        & High  & Low                   \\ \hline
In-Context                                                          & No                                                                   & GPU or GPU Cluster                                                              & Low   & High                   \\ \hline
\end{tabular}%
}
\caption{NLI classifiers are a good starting point for most tasks because they require no training, require relatively low compute, and are highly reproducible.}
\label{tab:class_types}
\end{table}

\textbf{Q1: Are your documents missing context necessary for classification?}\\
NLI classifiers are a good default approach for their accuracy, scalability, and reproducibility. However, they make inferences from a straightforward interpretation of documents and are unaware of context outside the text. For example, If a corpus contains documents about a politician not explicitly named, an NLI classifier is more likely to view those as irrelevant and thus, stance-neutral. Other examples may be a corpus with many documents that require abstract associations (e.g. Table \ref{tab:semeval}), or classification tasks using a specific operationalization of a concept (e.g. a particular definition of ``hate''). This can also include documents with coded, vague statements, or language requiring specific social knowledge to understand. For example, the phrase ``come and take it'' expresses support for gun ownership among American conservatives, but this cannot be inferred directly from the text.

There is no objective definition of sufficient context or how prominent ambiguous documents must be to warrant a different approach. Researchers must qualitatively judge based on how document relevancy was determined, and their familiarity with the data. However, because NLI classifiers are free and require no training, attempting one on a small test set and adjust your approach as necessary is low-commitment.

\noindent\textbf{Q2: Can you train a classifier for each stance I want to label?}\\
If an NLI classifier is insufficient, supervised and in-context classifiers both afford the researcher more control over model knowledge. Supervised classifiers offer better reproducibility and lower computational demands,  but are task-specific, potentially requiring multiple models for classifying multiple stance topics. Research on similar tasks suggest a class-balanced training set of 1,000-2,000 samples at minimum \citep{prabhu2021multi, margatina2021importance, laurer2024less}. If obtaining this many samples per topic is not feasible, consider an NLI or in-context classifier.

In-context classifiers may suit when other methods are infeasible. However, their utility is more limited due to model size and reproducibility concerns. While useful for small datasets, training data compilation, or validating other models, NLI or supervised classifiers are recommended when possible.

\subsection{Effects of Controlling Context}
\begin{figure}
    \centering
    \includegraphics[width = \textwidth]{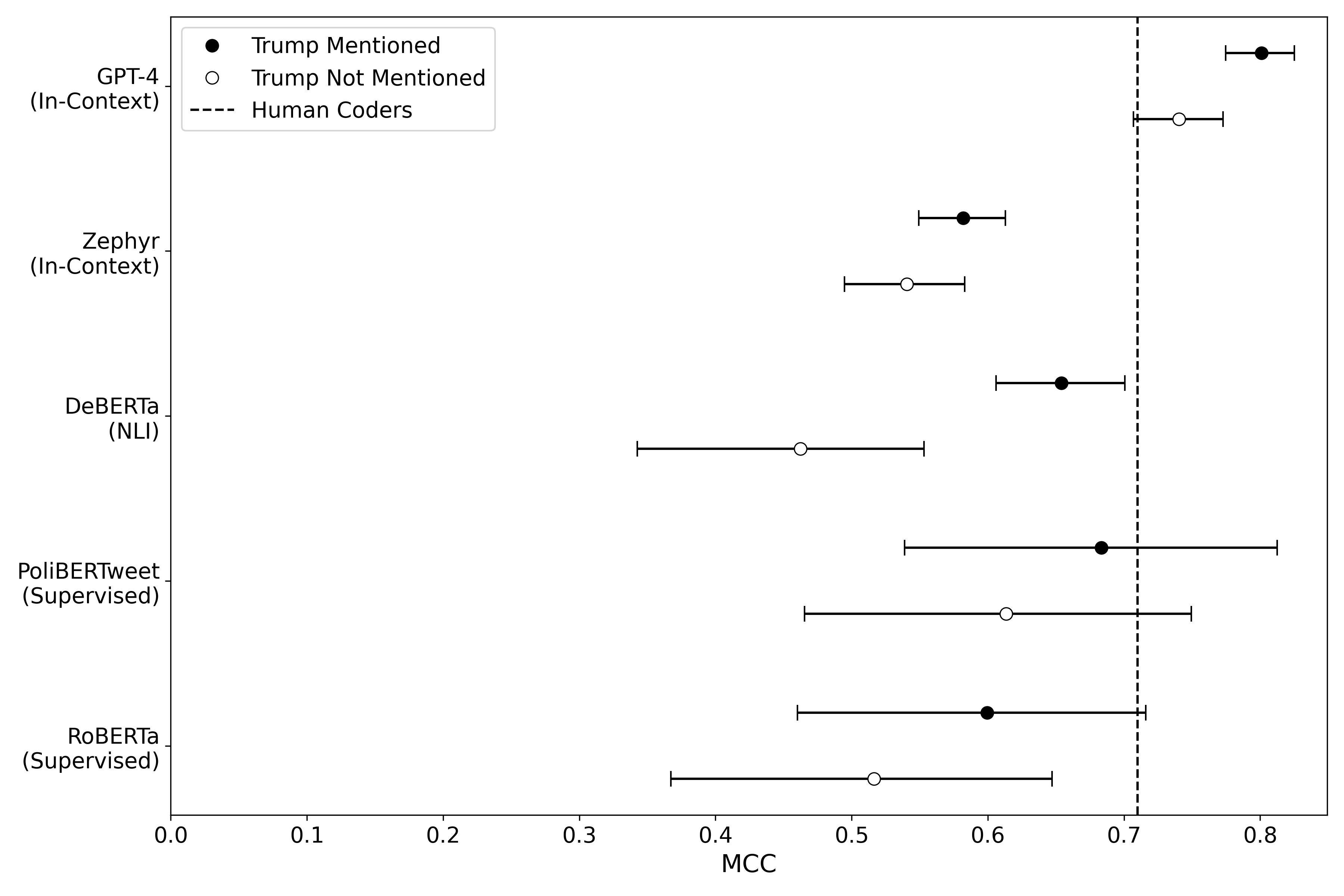}
    \caption{MCC on the testing data with bootstrapped 95\% confidence intervals. Considering what information your documents contain should inform your approach.}
    \label{fig:infocontext}
\end{figure}
To illustrate the impact of context, I categorized ``ambiguous'' documents in my test set as those with conflicting human coder assessments. I then assessed document relevance by the presence of the term ``Trump.'' Among documents mentioning Trump, 19.5\% were ambiguous. Conversely, 34.0\% of documents not mentioning Trump were ambiguous.\footnote{Using topic classification instead of keyword matching yielded similar results with 18.1\% and 32.6\% ambiguity, among documents that were and were not classified as Trump related respectively.} 

These differences in ambiguity directly affect classification performance. As shown in Figure \ref{fig:infocontext}, the NLI classifier suffers the most performance degradation when Trump is not mentioned. Supervised and in-context classifiers, however, adapt better to lower context. The implication is not to cull relevant observations due to ambiguity, but rather that researchers should consider what they know about documents in their corpus and try to address ambiguity with suitable classification methods.

\label{sec:frame}

\section{Approaches to Stance Detection}

\subsection{Supervised Classifiers}
Supervised classifiers learn patterns within a set of labeled training data and then identify those patterns in the broader corpus. The primary advantage of a supervised classifier is they afford researchers control over what the model ``knows'' via the training data. This makes them ideal for documents with low contextual information such as the examples in Table \ref{tab:semeval}. If a researcher knows that a pro-Univision stance implies an anti-Trump stance, they can teach this association to a supervised classifier through the training data. The primary weakness of supervised classifiers is that they are task specific, requiring additional training for new tasks. This makes them resource intensive for applications encompassing multiple topics or data sets. 

Because supervised classifiers are well established in the literature, I do not provide a comprehensive overview of their application here. Rather, I focus on selecting models and training data as these have stance detection specific considerations. For model training, I encourage adherence to best practices established elsewhere in the literature by \citet{terechshenko2020comparison}, \citet{laurer2024less}, \citet{ash2023text} and others.

\subsubsection{Model Selection}
Supervised classifiers can be divided into two types: bag-of-words models and language models. The bag-of-words approach uses word counts as features and an algorithm such as logistic regression or a random forest to label documents. These models offer advantages in interpretability and speed \citep{ash2023text}, but have relatively poor performance on entailment classification. Language models are neural networks pre-trained for general language comprehension, and then fine tuned to specific classification tasks -- a process known as transfer learning. They are computationally demanding, but are better entailment classifiers. 

Model choice has significant performance implications. Through a process called domain-adaptation, language models are pre-trained to learn language particular to a substantive domain such as legal language \citep{chalkidis2020legal} or social media text\citep{bertweet}. When there are no models adapted to a domain of interest, NLI classifiers can be re-trained as supervised classifiers for better performance with less data than other models \citep{laurer2024less}.

To demonstrate the significance of model selection, I trained a set of bag-of-words classifiers and language models. The language models all use the same neural network architecture but vary in their levels of domain adaptation. The first is a base RoBERTa \citep{liu2019roberta} model without domain adaptation, the second \citep[BERTweet;][]{bertweet} is adapted for Twitter, and the third \citep[PoliBERTweet;][]{kawintiranon2022polibertweet} is adapted for political Twitter.\footnote{Results in this section use Twitter data only.}

Table \ref{tab:sup_results} compares results across models. For each classifier I conducted a hyper-parameter sweep -- meaning I trained multiple instances of the model with different parameters to find the best combination. For bag-of-words classifiers I removed a standard set of English stop words, and vectorized documents using term-frequency-inverse-document-frequency and bi-grams. I used an exhaustive hyper-parameter sweep to find the optimal model. Because language models have many more hyper-parameters than other algorithms and take longer to train, an exhaustive sweep is infeasible. Rather, I trained each model ten times using a Bayesian algorithm to more efficiently sweep the parameter space. This approach first trains a model with randomized parameters and adjusts successive models based on results. Results in the table represent the best models. While compute times vary significantly based on data, hardware, and training procedure, the table provides estimates of relative training times in a single CPU or GPU context.

\begin{table}[]
\centering
\resizebox{\textwidth}{!}{%
\begin{tabular}{clrrrrr}
\cline{2-7}
\multicolumn{1}{l}{} & \multicolumn{1}{c}{\textbf{Model}} & \multicolumn{1}{c}{\textbf{MCC}} & \multicolumn{1}{c}{\textbf{F1}} & \multicolumn{1}{c}{\textbf{Accuracy}} & \multicolumn{1}{c}{\textbf{Sweep Time}} & \multicolumn{1}{c}{\textbf{Hardware}} \\ \cline{2-7} 
\multirow{3}{*}{\begin{tabular}[c]{@{}c@{}}Bag-of-Words\\ Classifiers\end{tabular}} & Logistic Regression & 0.51 & .67 & 77\% & \textbf{1.53s} & CPU \\
 & Random Forest & 0.49 & .62 & 76\% & 2 min. 2s & CPU \\
 & SVM & 0.49 & .68 & 76\% & 37.8s & CPU \\ \hline
\multirow{3}{*}{\begin{tabular}[c]{@{}c@{}}Language \\ Models\end{tabular}} & RoBERTa & 0.60 & .75 & 81\% & 46 min & GPU \\
 & BERTweet & 0.63 & .77 & 83\% & 44 min & GPU \\
 & PoliBERTweet & \textbf{0.68} & \textbf{.80} & \textbf{85\%} & 42 min & GPU\\ \hline
\end{tabular}%
}
\caption{Performance on detecting support for President Trump in Tweets. Language models offer better classification for higher compute demands. Bag-of-words classifiers were trained with an exhaustive grid search and a Bayesian sweep across 10 models was used to train the language models.}
\label{tab:sup_results}
\end{table}

A concern with language models among political scientists is the perception that compute demands are too high \citep{haffner2023introducing}. However, results here demonstrate models can quickly train on consumer grade GPUs -- such as those on free cloud computing platforms. Perhaps more significant than compute time is human labor. Here, too, language models have an advantage. \citet{laurer2024less} found that language models with 500 training examples consistently outperformed other algorithms with several thousand. This potentially reduces manual labeling dramatically. Further, converting documents to a bag-of-words requires text pre-processing that increases labor for researchers and alters results \citep{denny2018text}. This includes removing words assumed meaningless, removing suffixes, and dropping words that appear too frequently or infrequently. This necessitate evaluating how pre-processing affects results. Language models, however, minimize arbitrary decision points because they work with unedited documents. Thus, I recommend using language models for stance detection.

The results in Table \ref{tab:sup_results} also demonstrate the impact of domain adaptation. RoBERTa (no domain adaptation) performs the worst ($MCC = 0.60$), while BERTweet (adapted for Twitter) preforms slightly better ($MCC = 0.63$). PoliBERTweet (adapted for political speech on Twitter) preforms the best ($MCC = 0.68$). Additionally, I trained a base DeBERTaV3 model with no domain adaptation to test the significance of model architecture. DeBERTaV3 is an evolution of the BERT architecture that also uses improved training procedures. The model performed on par with BERTweet ($MCC = 0.63$) -- emphasizing that while more sophisticated models have advantages, domain adaptation should not be overlooked.

\subsubsection{Training Samples}
Training data controls the supervised classifiers knowledge base. Researchers should pay attention to how data is sampled and what manual labelers know. Discordance between labelers' knowledge and context required for accurate labeling limits model potential due to noisy training data \citep{miller2020active, joseph2017constance}. To ameliorate this, provide coders examples of correctly labeled text \citep{kawintiranon2021knowledge} or a prompt that provides necessary context \citep{joseph2017constance}. Crowd sourcing may not be appropriate if the context needed for accuracy is more than people can reasonably internalize in a short time. Here, training expert coders may be appropriate.

\subsubsection{Validation}
Validating classifiers is done by examining performance on data not seen during training. Two approaches are cross validation and using a test set. Cross validation randomly divides the data into several partitions. The model is trained on all data not in a particular partition, tested on the withheld data, then discarded and the process is repeated for all partitions \citep{cranmer2017can}. The results across partitions are pooled to estimate a generalized error rate.

A test set approach segments the dataset into a train, validation, and test set. The training set is used to train the model, the validation set monitors training progress, and the test set is withheld for testing after training completes \citep{cranmer2017can}. This approach offers efficiency advantages over cross validation because it only trains a single model. The train-validation-test split ratio depends on dataset size and desired error estimate precision. Common ratios are 70-15-15 and 60-20-20.

\subsection{NLI Classifiers}
NLI classifiers are language models pre-trained for recognizing textual entailment. Because most classification can be reformulated as an entialment task, NLI classifiers are ``universal,'' rather than task specific, classifiers. The process consists of pairing documents with hypothesis statements, and then assigning labels based on if a document entails a hypothesis \citep{yin2019benchmarking}. For example, I pair each document in my data with the hypothesis ``The author of this text supports Trump'' and the model returns a true or false classification.

\subsubsection{Model Selection}
Currently, the best language model for NLI classification is a DeBERTaV3 Large model trained on NLI datasets. It is the only model that achieves results comparable to humans, and is small enough for consumer hardware \citep{Superglue}. I recommend evaluating other models relative to this baseline. 

In evaluating models, first consider model size. As shown in Table \ref{tab:DeBERTa_size}, performance comparable to supervised classifiers only emerges in larger models. Domain adaptation or future advances may reduce the necessary model size. However, between models of the same architecture, larger models generally correlate with better performance.

\begin{table}[]
\centering
\resizebox{\textwidth}{!}{%
\begin{tabular}{lrrrrr}
\hline
\multicolumn{1}{c}{\textbf{Model}} & \multicolumn{1}{c}{\textbf{MCC}} & \multicolumn{1}{c}{\textbf{F1}} & \multicolumn{1}{c}{\textbf{Accuracy}} & \multicolumn{1}{c}{\textbf{\begin{tabular}[c]{@{}c@{}}Inference Time\\ (GPU)\end{tabular}}} & \multicolumn{1}{c}{\textbf{\begin{tabular}[c]{@{}c@{}}Inference Time\\ (CPU)\end{tabular}}} \\ \hline
DeBERTaV3 Small & 0.15 & 0.44 & 56\% & 3.06s & 2 min. 37s \\
DeBERTaV3 Base & 0.4 & 0.7 & 70\% & 4.98s & 5 min. 8s \\
DeBERTaV3 Large & \textbf{0.65} & \textbf{0.82} & \textbf{83\%} & 13.5s & 16 min. 35s \\ \hline
\end{tabular}%
}
\caption{Entailment classification without supervised training requires a relatively sophisticated language model.}
\label{tab:DeBERTa_size}
\end{table}

A second variable to consider is the training data. Stronger models are generally pre-trained on multiple large datasets curated for NLI, such as the ANLI \citep{nie2019adversarial} and WANLI \citep{liu2022wanli}. After pre-training, additional tuning on classification datasets can further improve performance. For example, the model trained by \citet{laurer2023building} was first pre-trained on multiple NLI data sets, and then tuned on a suite of smaller datasets to help it generalize across tasks.

\subsubsection{Hypotheses}
After selecting a model, a researcher pairs hypothesis statements with documents to classify them. This is generally done by creating a hypothesis template (e.g. ``The author of this text...'') and finishing the template with the relevant stance expression (e.g. ``...supports Trump.''). Appropriately pairing hypotheses with documents can dramatically impact performance. Consider the results in Figure \ref{fig:infocontext} that demonstrate a decline in performance when Trump is not mentioned. If a portion of our corpus refers to President Trump as ``Mr. President,'' we might change the hypotheses for those documents from ``The author of this text supports of Trump'' to ``The author of this text supports the President.''

Another consideration is how many hypotheses to use. NLI Classifiers can be paired with a single hypothesis or a set of hypotheses. In the single hypothesis instance, the model returns a probability of entailment. In the multi-label approach, each hypothesis is classified for entailment and the hypothesis most likely to be entailed is the assigned label. Results may differ slightly between approaches. Consider a multi-hypothesis scenario where the model chooses between ``support'', ``oppose'', and ``neutral.'' If the model finds that the entailment probability for each hypothesis is $<$ 0.5, it may still return the support label if it is the most likely among the three. In a single ``support'' hypothesis context, the document would be labeled as not supporting. The multi-label approach will increase computation time roughly linearly with the number of hypotheses since entailment probabilities are predicted for each label. It is recommended to try both approaches and validate results.

\subsubsection{Validation}
A straightforward approach to validation is to calculate the sample size needed to estimate performance within a confidence interval and label some data. Performance can be estimated with a 5-10\% margin of error at a 95\% confidence level with a fraction of the data needed to train a classifier. This can often be accomplished in a few hours.

Researchers should also conduct sensitivity analysis by generating synonymous hypotheses and classifying the data multiple times. This demonstrates results are not sensitive to hypothesis phrasing. For example, I classified my test set with 9 synonymous hypotheses (located in Appendix \ref{appendix: phrases}) for support of Trump. Across all hypotheses, the average MCC was 0.64, with a minimum value of 0.62 and a maximum value of 0.66. 

Classification results from another language model with a different architecture and training set can provide further validation. In-context learners, such as GPT-4, show particular promise here. The results on my test set shown in Figure \ref{fig:infocontext} and Table \ref{tab:prompts} demonstrate that GPT-4 out-preforms humans in identifying true labels. This could be used to expand the NLI classifier's test set, or as a second validation point against human labeled documents.

\subsection{In-Context Classification}
In-context learning refers to the capability of generative language models, such as GPT-4, to learn tasks via plain language prompts describing the task \citep{brown2020language}. Classification involves prompting the model with the task description, document, and a request for an appropriate label. Like supervised classifiers, they can make accurate inferences when the contextual information in documents is low. This is because both the model's pre-training and the user supplied prompt can provide information not contained in the document \citep{OpenAI_GPT4_2023}. Like NLI classifiers, their ability to label documents zero-shot means a single model can be used for many tasks. 

\subsubsection{Reproducibility Concerns}
Despite their advantages, in-context classifiers are not currently suitable for large-scale classification for two reasons: First, high performing models are currently proprietary models. Researchers cannot archive model versions for replication. Second, these models have costly compute requirements at scale. For reference, I include the cost of classifying my test set with the OpenAI API in Table \ref{tab:prompts}. My test set of 2,135 short documents cost \$12.52 to label with the cheapest approach. This cost quickly balloons as the number and length of documents increases. A data set of 1,000,000 Tweets would cost \$6,260 to label. Open source models that can run on local GPUs are available, but the cost is incurred via compute times. Zephyr 7B is an order of magnitude slower than an NLI classifier with no performance benefits.

Due to replication challenges, in-context classifiers have more limited application as classifiers. Where they show the most promise is in labeling smaller data sets, expanding training data for a supervised classifier, or validating other zero-shot classifiers alongside human labels. This technology will hopefully become more reproducible as it matures. Regardless, using non-reproducible models as the primary source of labels should be discouraged \citep{spirling2023open}.

\subsubsection{Model Selection}
Reproducibility, capability, cost, and speed are the factors to consider when selecting a model. While reproducibility is primarily a function of proprietary status, judging the other factors is less straight forward. I test three different models: GPT-4, GPT-3.5, and Zephyr 7B. GPT-4 is among the most capable models available across benchmarks \citep{OpenAI_GPT4_2023}. GPT-3.5 is a cheaper alternative that is comparable to other classification methods \citep{OpenAI_GPT4_2023}. Finally, Zephyr 7B is an open source model that can be archived and run on local computers for replication \citep{tunstall2023zephyr}.

To assess capability, popular models such as GPT-4 have been benchmarked for stance classification here and elsewhere \citep{tornberg2023chatgpt}. Absent direct testing, performance on shared benchmarks is informative. The HellaSwag benchmark \citep{zellers2019hellaswag}, for example, is commonly used to test language inference in generative models. Zephyr 7B and GPT3.5 score similarly on HellaSwag (84.52 and 85.5 respectively) and perform similarly in Table \ref{tab:prompts} \citep{OpenAI_GPT4_2023, tunstall2023zephyr}. GPT-4 scores substantially better (95.3) and stance detection reflects this \citep{OpenAI_GPT4_2023}.

Pricing for proprietary models is generally done per token, which are words or sub-words. The prompt, documents, and text generated by the model are all considered in calculating the total tokens. Estimating the cost of your data can be difficult because exact counts depend on the model used. OpenAI recommends assuming 100 tokens per 75 words \citep{OpenAI_tokens}.

Classification speed with proprietary models is determined by API rate limits and traffic.\footnote{The test set was classified multiple times via the OpenAI API. Times ranged from 35 minutes and 6.5 hours. Speed was not model dependent and entirely driven by API traffic.} With open source models, inference time varies based on document length, prompt length, and parameters used. At its fastest, Zephyr 7B took roughly 2.5 minutes to classify the test set on a consumer grade GPU. With a longer prompt and different parameters, the same task took over two hours.\\

\begin{minipage}{\linewidth}
\begin{lstlisting}[caption={An example prompt template in the the Python OpenAI package. The system message gives the model an ``identity'' and the user message explains the task and presents the document to be classified.}, captionpos=b, label={lst:prompt}]
system_message = "You are a text classifier and are only allowed to respond with 1 or 0."
user_message = "You are a classifier that determines if the author of a text supports Donald Trump. I will post text about Trump and if the author is more likely to support Trump than not return 1. If it is not more likely that the author supports Trump return 0. Do not explain the classification or say anything else. You are only allowed to respond with 1 or 0. Here is the text.\nText: {}"

messages = [
    {"role": "system", "content": system_message},
    {"role": "user", "content": user_message.format(document)}
]
\end{lstlisting}  
\end{minipage}

\subsubsection{Prompt Engineering}
Results vary based on the model prompt. Optimal prompt construction is an inexact science, but several validated approaches enchance in-context classification efficacy. Regardless of approach, prompt engineering should involve testing multiple prompts and validating results on a small labeled data set.

Prompts have two components: the system message and the user message. The system message instructs the model on how to behave and respond to prompts. The user message is the prompt the model responds to. User messages should explain contextual information necessary for accurate inference and present the document to be classified. The prompt should be formatted to distinctly mark the document requiring classification \citep{OpenAI_prompt}. Listing \ref{lst:prompt} provides an example prompt. The system message specifies the task and the range of desired responses. The user message provides context by specifying that the text concerns Donald Trump, reiterates the acceptable responses, and demarcates the text for classification.


One approach to prompting is few-shot learning. Few-shot prompts include example documents and their correct labels -- similar to training data. This approach has mixed results in improving classification \citep{brown2020language, kristensen2023chatbots}. Few-shot learning's challenge is generalizing from a small, potentially non-representative sample. This risks overfitting to prompted examples \citep{wang2020generalizing}. Thus, results can be highly variable depending on provided examples and arbitrary changes in the prompt \citep{perez2021true, lu2021fantastically}. These instabilities persist with more sophisticated models and more labeled examples \citep{zhao2021calibrate}. Thus, while few-shot prompts can improve classification in some cases, I recommend using them with caution -- particularly when zero-shot performance is sufficient.

Another prompting method is chain-of-thought reasoning \citep{wei2022chain}. This approach asks the model to explain its reasoning step-by-step before providing the conclusion. It improves performance on a variety of tasks, including classification, but not consistently so \citep{he2023annollm}. Chain-of-thought prompting increases costs significantly because the model generates an explanation in addition to the label. Evaluating chain-of-thought responses can be cumbersome because labels must be extracted from the rest of the response. Requesting models follow response templates with predictable label locations can help, though language models may inconsistently adhere to them. 

I evaluated chain-of-thought reasoning by constructing a prompt modeled after the one used by \citet{he2023annollm}. I found it offered no perceivable benefit in accuracy, but tripled the cost of GPT-4. I further found different compliance rates with the requested response template. GPT-4 showed 100\% compliance, while GPT-3.5 had 98.8\% compliance. Zephyr 7B uniformly failed template compliance and is thus not suitable for chain-of-thought classification.

\begin{table}[]
\centering
\resizebox{\textwidth}{!}{%
\begin{tabular}{clrrrcl}
\hline
\textbf{Prompt} & \multicolumn{1}{c}{\textbf{Model}} & \multicolumn{1}{c}{\textbf{MCC}} & \multicolumn{1}{c}{\textbf{F1}} & \multicolumn{1}{c}{\textbf{Accuracy}} & \textbf{\begin{tabular}[c]{@{}c@{}}Inference Time\\ (GPU)\end{tabular}} & \multicolumn{1}{c}{\textbf{Cost}} \\ \hline
\multirow{3}{*}{Standard} & GPT-4 & 0.78 & 0.88 & 89\% & - & \multicolumn{1}{r}{\$12.52} \\
 & GPT-3.5 Turbo & 0.58 & 0.73 & 81\% & - & \multicolumn{1}{r}{\$0.68} \\
 & Zephyr 7B & 0.59 & 0.74 & 79\% & \multicolumn{1}{r}{2 min. 25s} & \multicolumn{1}{c}{-} \\ \hline
\multirow{3}{*}{\begin{tabular}[c]{@{}c@{}}Chain\\ of\\ Thought\end{tabular}} & GPT-4 & 0.78 & 0.85 & 87\% & - & \$38.62 \\
 & GPT-3.5 Turbo & 0.56 & 0.69 & 77\% & - & \$1.26 \\
 & Zephyr 7B & \multicolumn{1}{c}{\xmark} & \multicolumn{1}{c}{\xmark} & \multicolumn{1}{c}{\phantom{blank}\xmark} & \multicolumn{1}{c}{\xmark} & \multicolumn{1}{c}{\xmark} \\ \hline
\multirow{3}{*}{Logit Bias} & GPT-4 & \textbf{0.8} & \textbf{0.89} & \textbf{90\%} & - & \$12.52 \\
 & GPT-3.5 Turbo & 0.63 & 0.81 & 81\% & - & \$0.68 \\
 & Zephyr 7B & 0.58 & 0.74 & 79\% & \multicolumn{1}{r}{4 min. 13s} & \multicolumn{1}{c}{-}
\end{tabular}%
}
\caption{Logit biasing can potentially improve performance for little to no additional cost. Chain-of-thought reasoning has shown mixed results on classification tasks but can significantly increase costs.}
\label{tab:prompts}
\end{table}

\subsubsection{Decoding Strategies}
The process of generating text from prompts is called decoding. There are many approaches models use, but in the context of classification the objective is to produce deterministic and reproducible results. This involves constraining the model's potential output tokens. There are three approaches to this: adjusting the temperature parameter, prompt construction, and biasing predicted token probabilities.

Temperature can be conceptualized as a randomization parameter. Generative models randomly sample the words they generate based on estimated probabilities \citep{welleck2020consistency}. The temperature parameter controls the shape of the distribution words are sampled from. A higher temperature flattens the distribution toward a uniform one where each word is equally likely as $T_{\to\infty}$ \citep{holtzman2020curious, ackley1985learning}. A temperature of zero ensures the most likely token is always chosen. Thus, I recommend setting temperature to zero.

Prompts can restrict the tokens models generate by specifying the desired response range. For example, I can instruct the model to only respond with ``1'' or ``0''. However, compliance will depend on model sophistication. GPT-4, for example, was 100\% compliant when asked to answer with only a ``1'' or ``0.'' Zephyr 7B showed less instruction-aligned results and required more prompt engineering for consistent binary labels on the test set.

To directly restrict the tokens models generate, users can manipulate token prediction scores, or ``logits.'' This approach applies a flat bias to the probabilities of tokens representing classes so that their predicted probability is much greater than other tokens. Table \ref{tab:prompts} shows that logit biasing generally has a positive effect on classification. Additionally, it simplifies label extraction by reducing non-compliant responses.

\subsubsection{Validation}
In-context classifiers should be validated with a test set. Calculate the number of samples needed to estimate performance within a desired confidence interval and use labeled data to test prompts and model parameters. Researchers should demonstrate that results are robust to synonymous prompts and not sensitive to arbitrary changes. Thus, researchers may want to classify their data with multiple prompts. An NLI classifier can also provide a second point of validation beside a sample of human labels. This demonstrates that two models with different architectures an training procedures obtain similar results.

\label{sec:guidance}

\section{Replication: COVID-19 Threat Minimization}

An analysis of Twitter posts by \citet{block2022perceived} showed that minimizing the threat of COVID-19 was in part ideologically motivated, but that the effect of ideology decreased as deaths within a person's geographic area increased. The original study used a supervised classifier to label documents. For this replication, I demonstrate how NLI classifiers can obtain similar results, as well as basic sensitivity analysis for zero-shot classification.

\subsection{Data and Design}
The replication data contains 862,923 tweets related to COVID-19 posted between September 1, 2020 and February 28, 2021 from 23,476 unique users. Each user's ideology is measured on a uni-dimensional left-right scale using tweetscores, a network based ideal point estimation \citep{barbera2015birds}. Included are 2,000 hand labeled training tweets and coding rules for labeling tweets ``non-compliant.'' Non-compliant tweets include statements that downplay the threat of COVID-19 (e.g. comparing it to the flu), as well as statements against mitigation practices (e.g. vaccination).

Based on the coding rules, I created two sets of synonymous entailment hypotheses with compliant, non-compliant, and neutral stance statements. A complete set of hypotheses is listed in Appendix \ref{appendix: phrases}. I then used keywords to match hypotheses to tweets and classified the remaining corpus with a DeBERTaV3 NLI classifier. If a tweet entails any of the threat minimizing hypotheses it is considered threat minimizing. For example, a tweet that contains the words ``mask'' and ``vaccine'' will be classified once for the hypotheses associated with masks and again for hypotheses associated with vaccines. If the model determines the document entails an anti-mask or anti-vaccine hypothesis it is considered threat minimizing. I use 300 randomly sampled tweets from the training data to test the hypotheses -- enough data to estimate performance at a 95\% confidence level with a <5\% margin of error, and a small enough sample that a single researcher could label the data in a few hours.

To analyze results I replicate the model used by \citet{block2022perceived} -- a negative binomial regression with a count of threat-minimizing tweets made as the dependent variable. The independent variables are the user's ideology, the COVID-19 death rate within their county, and an interaction between the two. Control variables including county level demographics, political lean, and state fixed effects are also used. 

\subsection{Analysis}
 
\begin{figure}
    \centering
    \includegraphics[width=\textwidth]{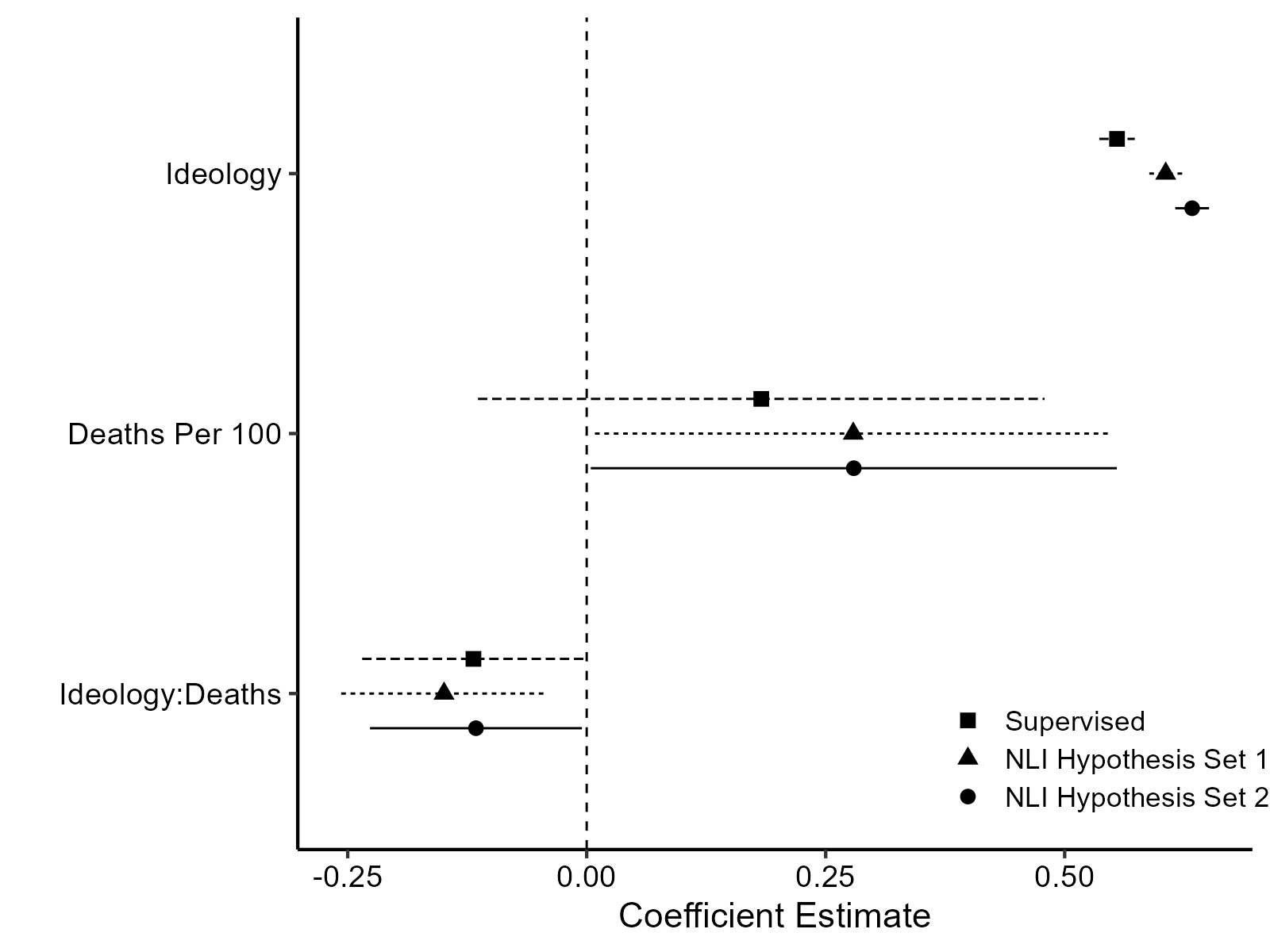}
    \caption{Replication results from \citep{block2022perceived}. Supervised represents the original results from a trained Electram model, while the NLI hypothesis sets used a DeBERTaV3 model for zero-shot classification.}
    \label{fig:dotwhisker}
\end{figure}

Using the NLI classifier on my sample of 300 labeled tweets, the first set of hypotheses had an MCC of 0.70 and 89\% accuracy while the second had an MCC of 0.75 and 91\% accuracy. The original classifier achieved an MCC of 0.66 and accuracy of 86\% on the test set. Between the models there was a high degree of intercoder reliability ($MCC = 0.84$). Figure \ref{fig:dotwhisker} shows the results from regression models that use zero-shot labels from both sets of hypotheses, as well as the original labels from the supervised classifier. The three models show consistency in both the direction and size of the effect.

\begin{figure}
    \centering
    \includegraphics[width=\textwidth]{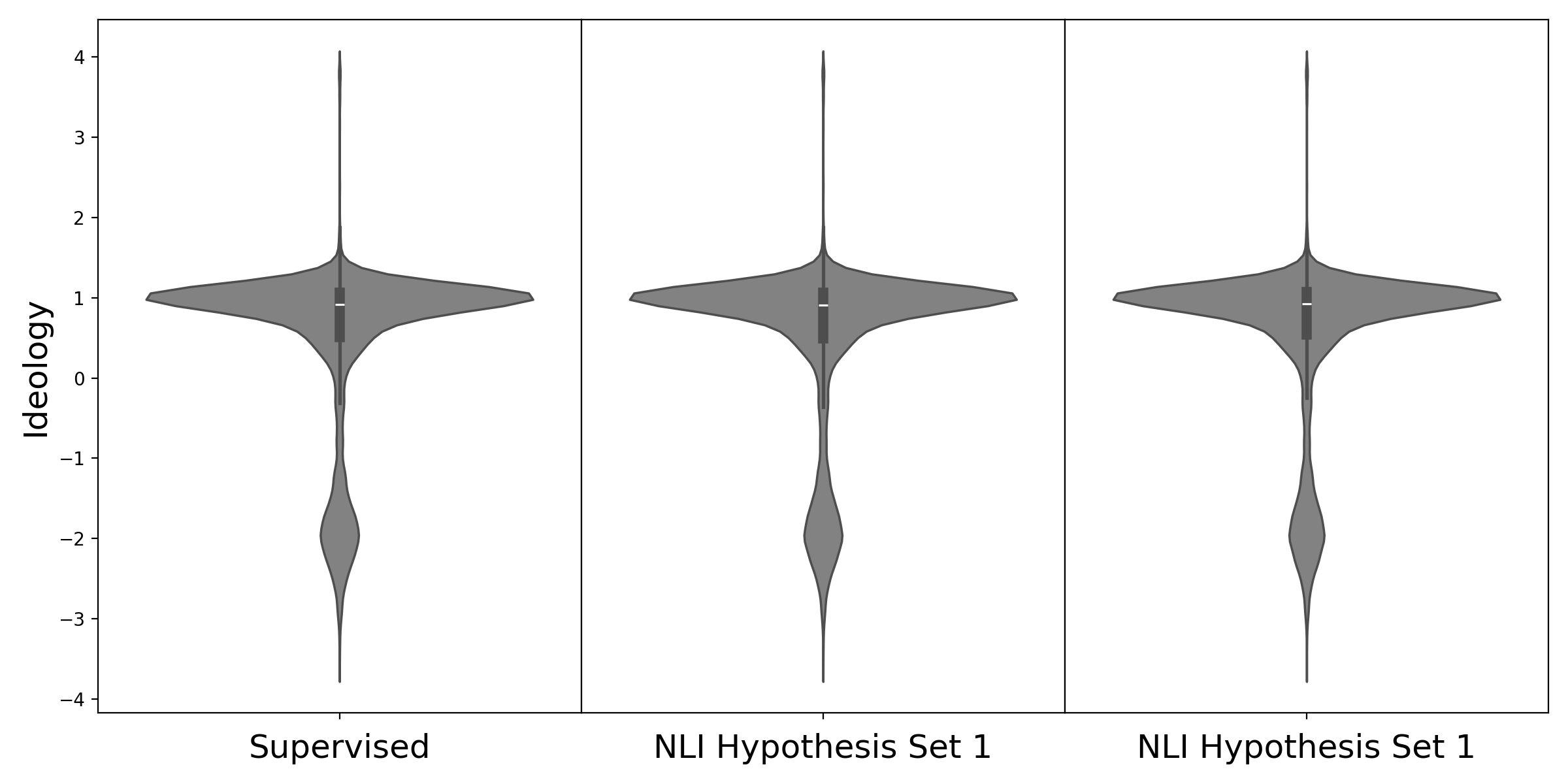}
    \caption{The distribution of tweet author ideology among tweets labeled threat minimizing by the three classifiers.}
    \label{fig:ideal_dist}
\end{figure}

Figure \ref{fig:ideal_dist} shows the ideological distribution of all tweets labeled threat-minimizing across the three models. The distributions appear identical with each identifying a concentration of threat minimizing tweets on the conservative pole. 
Across the entire sample, the average ideology of a tweet's author is -0.66. Among samples labeled differently by the zero-shot classifier and the original classifier, the average ideology was -0.11 for the first set of hypotheses, and -0.07 for the second set of hypotheses -- indicating the NLI classifiers are slightly more likely to disagree with the supervised model when the author is more conservative. However, the models show a high level of consistency across the data in identifying a relatively nuanced ``threat minimization'' stance. This provides compelling evidence that zero-shot NLI models can be a useful avenue for stance detection.

\label{sec:ex2}

\section{Conclusion}

In this paper, I outlined a precise definition and generalized framework for stance detection. Stance detection is entailment classification and determines how a document responds to a proposition. After defining stance proposition, there are currently three possible classification approaches: supervised classification, NLI classification, and in-context classification. Researchers have two primary considerations to determine the right approach: What information is necessary for accurate classifications, and how to balance trade-offs between resources, computing demand, and human labor. When the information documents contain is sufficient for accurate classification, NLI classifiers achieve performance comparable to supervised classifiers while eliminating the need to train a model. When context outside of the document is necessary for classification, supervised and in-context classifiers are a better choice.

A potentially fruitful direction for future research is in adapting NLI classifiers to better understand political text. Advancements in NLI classification are largely from computer science, and the data models are trained on reflects this. There are no NLI data sets specifically for stance detection, or stance detection benchmarks. Data sets and pre-trained models with political communication in mind could mean more reliable zero-shot classification with more accessible compute demands.

Finally, stance detection methods can be advanced by moving beyond classification and towards scaling the intensity of a stance. This could possibly be achieved by integrating sentiment measures, but more research is needed.

\label{sec:conclusion}

\clearpage








\singlespacing
\bibliographystyle{te}

\bibliography{refs}

\clearpage
 \appendix
 \section{Method Summaries}
 \label{summaries}

\begin{singlespacing}
\begin{center}
\fbox{\begin{minipage}{\textwidth}
\begin{center} \textbf{Recommendations for Supervised Classifiers}\end{center}
\textbf{Selecting a Model}
\begin{itemize}
    \item Language models provide superior classification over bag-of-words classifiers.
    \item Use a domain adapted model if one is available. Otherwise, models trained for NLI learn faster than base models.
\end{itemize}

\textbf{Training Samples}
\begin{itemize}
    \item Ensure manual labelers have sufficient context to accurately label documents.
    \item Train expert coders if your classification task requires domain specific expertise.
\end{itemize}

\textbf{Validation}
\begin{itemize}
    \item Out of sample prediction is the primary test of validity.
    \item Cross validation can be computationally expensive for language models, so most applications opt to use a test set.
\end{itemize}
\end{minipage}}
\end{center}
\end{singlespacing}

\begin{singlespacing}
\begin{center}
\fbox{\begin{minipage}{\textwidth}
\begin{center} \textbf{Recommendations for NLI Classifiers}\end{center}

\textbf{Selecting a Model}
\begin{itemize}
    \item Use a model pre-trained on multiple NLI data sets.
    \item DeBERTaV3 Large is currently the only model to achieve performance comparable to supervised classifiers on entailment classification. Use this as a baseline for comparing other models.
\end{itemize}

\textbf{Generating Hypotheses}
\begin{itemize}
    \item Match hypotheses to documents based on some relevancy parameter such as keyword matching or topic classification.
    \item Try classification with a single and multi-hypothesis approach.
\end{itemize}

\textbf{Validation}
\begin{itemize}
    \item Calculate the number of samples needed to estimate performance within a confidence interval and label some data.
    \item Conduct basic sensitivity analysis by classify documents multiple times with synonymous hypotheses.
    \item Consider using an in-context classifier like GPT-4 as a second point of validating labels.
\end{itemize}
\end{minipage}}
\end{center}
\end{singlespacing}

\begin{singlespacing}
\begin{center}
\fbox{\begin{minipage}{\textwidth}
\begin{center} \textbf{Recommendations for In-Context Classifiers}\end{center}

\textbf{Model Selection}
\begin{itemize}
    \item When using an in-context classifier to label training or validation data, prioritize model capability. Currently, GPT-4 and Claude 3 Opus are the most sophisticated models.
    \item When using an in-context classifier as the the primary classifier, use an open source and reproducible model.
    \item HuggingFace.co hosts a leaderboard for comparing open source models. The HellaSwag benchmark is useful for assessing a model's inference capabilities.
\end{itemize}

\textbf{Prompt Engineering}
\begin{itemize}
    \item Use the system message to prime the model's behavior and the user message to describe the classification task.
    \item Few-shot classification is inadvisable if zero-shot classification is sufficient due to the potential for over-fitting to prompt examples.
    \item Chain-of-thought reasoning has inconsistent effects, but significantly increases costs.
\end{itemize}

\textbf{Parameters and Decoding}
\begin{itemize}
    \item Turn the temperature parameter to zero.
    \item Restrict the possible tokens a model can generate to those that represent your classes by biasing the logits.
\end{itemize}

\textbf{Validation}
\begin{itemize}
    \item Estimate performance within a margin of error on a small sample of labeled data.
    \item Conduct sensitivity analysis by classifying the data multiple times with synonymous prompts and showing results are robust across prompts.
\end{itemize}
\end{minipage}}
\end{center}
\end{singlespacing}

 \clearpage


 \section{Testing Data and Entailment Phrases}

\subsection{Testing Data}
\begin{table}[!h]
    \centering
    \begin{tabular}{{llcp{5cm}}}
        \hline
        Data set & Source & N & Notes \\
        \hline
        MTSD & \cite{sobhani2017dataset}  &  398 & Tweets. Each tweet contains multiple politicians. No context provided to labelers. \\
        
        ConStance & \cite{joseph2017constance}   & 98& Tweets. Annotators given a contextual prompt. Assumes anti-Clinton statements are pro-Trump.  \\
        
        KE-MLM & \cite{kawintiranon2021knowledge}  &  380& Tweets. Annotators given examples of tweets with appropriate labels.  \\
        
        SemEval2016 & \cite{mohammad2016semeval}  &  97 & Tweets. No context provided to labelers.  \\
        
        Titling & \cite{van2019not}  &  161 & Tweets. Annotators given the author's photo, location, description, and text the tweet was responding to.  \\
        Newsletters & \cite{cormack2017dcinbox} & 1,000 & Newsletter Sentences. No context provided to labelers.\\
        \hline
    \end{tabular}
    \caption{Data sources and the number of observations per target}
    \label{tab:ex1_sources}
\end{table}

\subsection{Trump Approval Stance Hypotheses}
\subsubsection{Default Hypothesis}
The author of this text supports Trump.
\subsubsection{Alternative Hypotheses}
The author of this text is pro Trump.\\
The author of this text approves of Trump.\\
The person who wrote this text supports Trump.\\
The person who wrote this text is pro Trump.\\
The person who wrote this text approves of Trump.\\
The author of this document supports Trump.\\
The author of this document is pro Trump.\\
The author of this document approves of Trump.\\

\begin{table}[]
\centering
\resizebox{\textwidth}{!}{%
\begin{tabular}{|l|l|l|}
\hline
\multicolumn{1}{|c|}{\textbf{Dimension}} & \multicolumn{1}{c|}{\textbf{Non-compliant}} & \multicolumn{1}{c|}{\textbf{Compliant}} \\ \hline
Death & \begin{tabular}[c]{@{}l@{}}The author of this tweet  believes \\ covid has not increased deaths.\end{tabular} & \begin{tabular}[c]{@{}l@{}}The author of this tweet believes \\ covid has increased deaths.\\ The author of this tweet believes \\ covid deaths are neutral.\end{tabular} \\ \hline
Flu & \begin{tabular}[c]{@{}l@{}}The author of this tweet believes\\ covid is like the flu.\end{tabular} & \begin{tabular}[c]{@{}l@{}}The author of this tweet believes\\ covid is not like the flu.\\ The author of this tweet believes\\ nothing about covid and the flu.\end{tabular} \\ \hline
Masks & \begin{tabular}[c]{@{}l@{}}The author of this tweet believes\\ masks are bad.\end{tabular} & \begin{tabular}[c]{@{}l@{}}The author of this tweet believes\\ masks are good.\\ The author of this tweet believes\\ masks are neutral.\end{tabular} \\ \hline
Lockdowns & \begin{tabular}[c]{@{}l@{}}The author of this tweet believes\\ lockdowns are bad.\end{tabular} & \begin{tabular}[c]{@{}l@{}}The author of this tweet believes\\ lockdowns are good.\\ The author of this tweet believes\\ lockdowns are neutral.\end{tabular} \\ \hline
Vaccines & \begin{tabular}[c]{@{}l@{}}The author of this tweet believes\\ vaccines are bad.\end{tabular} & \begin{tabular}[c]{@{}l@{}}The author of this tweet believes\\ vaccines are good.\\ The author of this tweet believes\\ vaccines are neutral.\end{tabular} \\ \hline
\begin{tabular}[c]{@{}l@{}}Social \\ Distancing\end{tabular} & \begin{tabular}[c]{@{}l@{}}The author of this tweet believes\\ social distancing is bad.\end{tabular} & \begin{tabular}[c]{@{}l@{}}The author of this tweet believes\\ social distancing is good.\\ The author of this tweet believes\\ social distancing is neutral.\end{tabular} \\ \hline
\begin{tabular}[c]{@{}l@{}}COVID-19\\ General\end{tabular} & \begin{tabular}[c]{@{}l@{}}The author of this tweet believes\\ the pandemic is not dangerous.\\ The author of this tweet believes\\ covid is not dangerous.\\ The author of this tweet believes\\ coronavirus is not dangerous.\end{tabular} & \begin{tabular}[c]{@{}l@{}}The author of this tweet believes\\ the pandemic is dangerous.\\ The author of this tweet believes\\ the pandemic is neutral.\\ The author of this tweet believes\\ covid is dangerous.\\ The author of this tweet believes\\ covid is neutral.\\ The author of this tweet believes\\ coronavirus is dangerous.\\ The author of this tweet believes\\ coronavirus is neutral.\end{tabular} \\ \hline
\end{tabular}%
}
\end{table}

\begin{table}[]
\centering
\resizebox{\textwidth}{!}{%
\begin{tabular}{|l|l|l|}
\hline
\multicolumn{1}{|c|}{\textbf{Dimension}} & \multicolumn{1}{c|}{\textbf{Non-compliant}} & \multicolumn{1}{c|}{\textbf{Compliant}} \\ \hline
Death & \begin{tabular}[c]{@{}l@{}}The author of this tweet believes\\ COVID death counts are wrong.\end{tabular} & \begin{tabular}[c]{@{}l@{}}The author of this tweet believes\\ many people have died from COVID.\\ The author of this tweet does not\\ express an opinion about COVID deaths.\end{tabular} \\ \hline
Flu & \begin{tabular}[c]{@{}l@{}}The author of this tweet believes\\ COVID is similar to the flu.\end{tabular} & \begin{tabular}[c]{@{}l@{}}The author of this tweet does not\\ believe COVID is similar to the flu.\\ The author of this tweet does not\\ compare COVID and the flu.\end{tabular} \\ \hline
Masks & \begin{tabular}[c]{@{}l@{}}The author of this tweet opposes\\ wearing masks.\end{tabular} & \begin{tabular}[c]{@{}l@{}}The author of this tweet supports\\ wearing masks.\\ The author of this tweet does not\\ express an opinion about wearing masks.\end{tabular} \\ \hline
Lockdowns & \begin{tabular}[c]{@{}l@{}}The author of this tweet opposes\\ lockdowns.\end{tabular} & \begin{tabular}[c]{@{}l@{}}The author of this tweet supports\\ lockdowns.\\ The author of this tweet believes\\ lockdowns are bad for the economy.\\ The author of this tweet believes\\ lockdowns save lives.\\ The author of this tweet does not\\ express an opinion about lockdowns.\end{tabular} \\ \hline
Vaccines & \begin{tabular}[c]{@{}l@{}}The author of this tweet opposes\\ vaccines.\end{tabular} & \begin{tabular}[c]{@{}l@{}}The author of this tweet supports\\ vaccines.\\ The author of this tweet believes\\ vaccines are dangerous.\\ The author of this tweet believes\\ vaccines are safe.\\ The author of this tweet does not\\ express an opinion about vaccines.\end{tabular} \\ \hline
\begin{tabular}[c]{@{}l@{}}Social \\ Distancing\end{tabular} & \begin{tabular}[c]{@{}l@{}}The author of this tweet opposes\\ social distancing.\end{tabular} & \begin{tabular}[c]{@{}l@{}}The author of this tweet supports\\ social distancing.\\ The author of this tweet does not\\ express an opinion about social distancing.\end{tabular} \\ \hline
\begin{tabular}[c]{@{}l@{}}COVID-19\\ General\end{tabular} & \begin{tabular}[c]{@{}l@{}}The author of this tweet does not\\ believe the pandemic is dangerous.\\ The author of this tweet does not\\ believe COVID is dangerous.\\ The author of this tweet does not\\ believe Coronavirus is dangerous.\end{tabular} & \begin{tabular}[c]{@{}l@{}}The author of this tweet believes\\ the pandemic is dangerous.\\ The author of this tweet believes\\ the pandemic is neutral.\\ The author of this tweet believes\\ COVID is dangerous.\\ The author of this tweet believes\\ COVID is neutral.\\ The author of this tweet believes\\ Coronavirus is dangerous.\\ The author of this tweet believes\\ Coronavirus is neutral.\\ The author of this tweet does not\\ express an opinion about the pandemic.\\ The author of this tweet does not\\ express an opinion about Coronavirus.\\ The author of this tweet does not\\ express an opinion about COVID.\end{tabular} \\ \hline
\end{tabular}%
}
\end{table}

 \label{appendix: phrases}
 \end{document}


\subsection{Definitions}

\textbf{Batch Size:} The number of training examples used in one iteration or weight update during training. Higher batch sizes require more memory usage but increase computational efficiency.

\textbf{BERT (Bidirectional Encoder Representations from Transformers):} A language model that processes input sequences (text) in both directions, allowing it to understand the context of a word based on both its preceding and following words. This bidirectional approach overcame the limitations of previous unidirectional models and significantly improved the performance of various natural language processing tasks.

\textbf{Chain-of-Thought Reasoning:} A method of prompting generative language models that requests they explain their reasoning before providing a conclusion. This can sometimes improve model performance.

\textbf{Context:} Information relevant to the stance of interest. Available context consists of information the document contains, and the knowledge base of the classifier. Missing context increases document ambiguity.

\textbf{CPU:} Central Processing Unit. The primary component of a computer responsible for executing instructions from software. The CPU is generally too slow for training and running large language models.

\textbf{DeBERTa (Decoding-enhanced BERT with Disentangled Attention):} Another extension of BERT that introduces two novel techniques: disentangled attention and enhanced mask decoder. Disentangled attention separates information about a tokens content and sentence position, allowing the model to capture richer semantic patterns. The enhanced mask decoder improves the model's ability to reconstruct masked tokens, leading to better understanding of natural language.

\textbf{DeBERTaV3:} DeBERTaV3 uses a more advanced pre-training approach to improve upon DeBERTa. DeBERTaV3 has demonstrates state-of-the-art NLI classification among models similar models such as BERT, RoBERTa, and Electra.

\textbf{Decoding:} The process used to create output sequences from a generative language model.

\textbf{Domain Adaptation:} The process of adjusting a model trained on one domain to perform well on a different domain.

\textbf{Electra:} An evolution of BERT that improves upon the pre-training methods used. It trains two transformer models simultaneously: a generator that predicts masked tokens, and a discriminator that distinguishes the generated tokens from the original ones.

\textbf{Few-shot Classification:} A setting where a language model is provided with a small number of labeled examples for each category and is expected to generalize to classify new instances accurately.

\textbf{Generative Language Models:} Language models that are trained to generate human-like text by predicting the probability distribution of the next word or token based on the previous context. Examples include GPT-4, Claude 3, and Llama 2.

\textbf{Generative Pretrained Transformer (GPT):} A type of transformer-based language model that is pretrained on a large corpus of text to predict the next token in a sequence. This allows it to generate human-like text.

\textbf{GPU:} Graphics Processing Unit. A specialized computer component designed to accelerate the processing of images and parallel computations. GPUs are frequently used to train language models.

\textbf{Hyperparameter:} Settings that control the behavior of a machine learning algorithm or language model. These include such as learning rate, batch size, or training epochs.

\textbf{Hyperparameter Sweep:} Systematically testing different combinations of hyperparameters to find the optimal configuration for a task or dataset.

\textbf{Hypothesis:} the proposition on which an author may take a stance. A stance statement.

\textbf{In-context Learning:} A paradigm where a language model learns to perform a new task by describing the task in a prompt, without explicit fine-tuning.

\textbf{Learning Rate:} A hypterparameter determining how much the model's weights are updated during training iterations. Higher rates allow faster convergence but risk overshooting the optimal solution. Lower rates risk getting trapped in locally optimal solutions and missing globally optimal solutions.

\textbf{Logit Bias:} A technique used to adjust the output distribution of a language model by adding a bias term to the logits (pre-normalized output scores). This allows for control over the model's generation or classification behavior.

\textbf{Natural Language Inference (NLI):} The task of determining the logical relationship between a pair of documents (entailment, contradiction, or neutrality). Models that are effective at NLI can be used as universal classifiers.

\textbf{Prompt:} A plain language input sequence provided to a generative language model to elicit a desired response.

\textbf{Prompt Engineering:} The process of designing and refining prompts to optimize the performance of generative language models on specific tasks.

\textbf{RoBERTa (Robustly Optimized BERT Pretraining Approach):} An improved version of BERT. It uses the same model as BERT but addresses some of its shortcomings by employing different pre-training strategies and more training data.

\textbf{Sentiment:} Positive or negative emotional valence.

\textbf{Stance:} How an individual would answer a proposition.

\textbf{Stance Detection:} Text sample \textit{T} entails stance \textit{S} to author \textit{A} when a human reading \textit{T} with context \textit{C} would infer that \textit{T} expresses support for \textit{S}.

\textbf{Supervised Classification:} A machine learning task where the model is trained on labeled data to classify new, unseen instances into predefined categories. Supervised classifiers are task specific.

\textbf{System Message:} A special type of prompt used in conversational AI systems to set the tone, persona, or expected behavior of the language model during the conversation.

\textbf{Temperature:} A hyperparameter that controls the randomness of the output from a generative language model. A temperature of zero leads to the most deterministic outputs.

\textbf{Textual Entailment:} Text sample \textit{T} entails hypothesis \textit{H} when a human reading \textit{T} would infer that \textit{H} is most likely true.

\textbf{Token:} A token is the basic unit of input and output in natural language processing models. It can be a word, sub-word, character, or any other meaningful unit that the model operates on. Text data is typically tokenized (split into tokens) before being fed into the model.

\textbf{Training Epochs:} One complete pass through the entire training dataset. Multiple epochs are typically required for the model to converge and achieve optimal performance.

\textbf{Transfer learning:} A machine learning technique where a model pre-trained on one task then adapted or fine-tuned for another related task.

\textbf{Transformers:} A type of neural network architecture that employs self-attention mechanisms to capture long-range dependencies between words in sequences. Transformers consist of an encoder portion that creates semantic representations of documents, and a decoder portion that uses these representations to generate text. Transformers are the the backbone of many modern language models from BERT to GPT-4. BERT type models primarily use the encoder portion, while GPT type models primarily use the decoder.

\textbf{Zero-shot Classification:} The ability of a language model to classify documents into categories it has not been explicitly trained on. NLI and in-context classifiers are capable of zero-shot classification.


This appendix provides basic guidance on hardware requirements for using language models, where to find language models, as well as software package recommendations. 

\subsection{Hardware}
While hardware requirements will vary based on the selected classification approach, there are a few considerations and resources common across methods. First, researchers will likely want access to an Nvidia GPU. Stance detection is best done with large language models and both training and inference can be prohibitively time consuming without a GPU. Nvidia GPUs are required because almost all contemporary deep learning libraries are developed in CUDA, a parallel computing platform developed by Nvidia for Nvidia GPUs. Compute clusters and cloud GPU services will likely only offer access to Nvidia GPUs. 

In most instances, a single consumer grade GPU such as those found in desktop computers is sufficient. If local access to a GPU is not available, Google Colab is likely the easiest way to get GPU access. Colab's free tier offers GPU access to anyone with a google drive based on server traffic. Long run times on the free tier will likely be interrupted, but free access is usually sufficient to train models or test your code before pushing it to a paid server or a university compute cluster. Kaggle (\url{https://www.kaggle.com/code}) also offers limited free GPU access appropriate for supervised and NLI classification.

\subsection{Language Models}
Open source language models are generally released on the Hugging Face Hub (\url{huggingface.co/models}), an online platform that hosts hundreds of thousands of models as well as tens of thousands of data sets. The hub hosts models for supervised training, zero-shot NLI classification, as well as in-context classifiers. Models generally include information about the data they were trained on, performance benchmarks, and code templates for further training or implementing the model. Hugging Face also host a leaderboard (\url{https://huggingface.co/spaces/HuggingFaceH4/open_llm_leaderboard}) for open source generative language models that can aid in model selection.

The list of proprietary models that can be used is regularly expanding. These include the GPT models developed by OpenAI and tested in this paper, as well as models trained by other AI companies such as Google and Anthropic. Currently, OpenAI has the most accessible API with the most well developed software package and documentation.

\subsection{For Python Users}
The Transformers library is the most popular open source library for using large language models. The package is maintained by Hugging Face, Inc. and is integrated with the model hosting hub as well as other libraries for managing data sets and models. The library is built on top of the PyTorch or Tensorflow deep learning libraries, with PyTorch being the most widely adopted. Third party libraries such as SimpleTransformers and SentenceTransformers provide alternative or task-specific libraries that are integrated with models hosted on the Hugging Face Hub. For individuals looking to get started with language models, I recommend installing the PyTorch version of the Transformers library.

Many companies that host proprietary language models also maintain Python packages for interfacing with the models. OpenAI maintains a packaged with thorough documentation and examples on the OpenAI website. Anthroptic and Google also maintain packages for interfacing with their Claude and PaLM models.

\subsection{For R Users}
Resources for using language models is relatively limited in R compared to Python. Natural language processing libraries are primarily developed in Python and do not currently have well developed counterparts in R. R users may find less operability with GPUs as CUDA is much more widely used via Python than R. Proprietary models, such as GPT-4, do not have R libraries maintained by the hosting companies either. If possible, it is recommended to simply use Python as there will be more helpful resources and documentation online for troubleshooting. This will likely be the easiest approach in the long run even if you have never used Python.

Perhaps the most common method of using large language models with R is simply using the Python transformers library through R's reticulate package. The recently released Text library provides some functionality for R users wanting to train large language models and use the Hugging Face hub, but is relatively limited compared to the Transformers library \citep{textpackage}.

\subsection{Memory Management}
When using a large language model, it is loaded on to a GPU to preform its computations. The amount of memory the GPU has will be the primary limitation when working with language models as large models can quickly occupy the entirety of the available memory. Generally, there are two different states of model use with very different memory requirements: training, and inference. Training adjusts the waits of the model to learn a task from training data. This process is the more memory intensive process of the two. Inference is when data makes a forward pass through the model and the network produce an output. In both cases, more memory is consumed when longer or multiple documents or prompts are passed through the model.

The most straightforward way to manage memory is by using multiple GPUs to increase to pool of available memory. However, absent that, there are other approaches to significantly reduce memory requirements. Batch size refers to the number of documents passed through the model at once. Higher batch sizes speeds up training and inference time but also increases memory demands. A batch size of 16 is a good starting point for single GPU applications. Lower or increase batch sizes as needed.

Quantized models are models in which the decimal precision of weights in the model have been reduced \citep{frantar2022gptq}. This significantly reduces the memory requirements of the model and makes calculations faster. Often, models can be quantized with minimal impact on performance. Quantized versions of larger models can often be found on the Hugging Face hub. If none are available, the Optimum library maintained by Hugging Face provides a simple pipeline for quantizing models.

LoRA adaptation stands for Low-Rank Adaptation of Large Language Models. It is a method model training that dramatically reduces the number of parameters that are adjusted, and thus the memory requirements \citep{hu2021lora}. This may lower performance in some instances, but can often be done with no cost to performance. The PEFT library maintained by Hugging Face offers accessible pipelines for LoRA tuning.